\documentclass[conference]{IEEEtran}
\IEEEoverridecommandlockouts
\usepackage{cite}
\usepackage{amsmath,amssymb,amsfonts}
\usepackage{algorithmic}
\usepackage{graphicx}
\usepackage{textcomp}
\usepackage{xcolor}
\usepackage{pifont}
\usepackage{url}
\usepackage[caption=false]{subfig}
\def\BibTeX{{\rm B\kern-.05em{\sc i\kern-.025em b}\kern-.08em
    T\kern-.1667em\lower.7ex\hbox{E}\kern-.125emX}}
\begin{document}

\title{ParkingScenes: A Structured Dataset for End-to-End Autonomous Parking in Simulation Scenes\\
\thanks{This work was supported by the National Key Research and Development Program of China (2022YFB4703700), the Key-Area Research and Development Program of Guangdong Province (2020B0909050001), the Key Research and Development Program of Shaanxi Province (2024CY2-GJHX-49). 

\emph{* Corresponding author.}}
}

\author{
\IEEEauthorblockN{1\textsuperscript{st} Haonan Chen}
\IEEEauthorblockA{\textit{Institute of Automation} \\
\textit{Chinese Academy of Sciences}\\
Beijing, China \\
chenhaonan2024@ia.ac.cn}

\and
\IEEEauthorblockN{2\textsuperscript{nd} Kaiwen Xiao}
\IEEEauthorblockA{\textit{School of Art and Design} \\
\textit{Yanshan University}\\
Hebei, China \\
13718031392@163.com}

\and
\IEEEauthorblockN{3\textsuperscript{rd} Bin Tian*}
\IEEEauthorblockA{\textit{Institute of Automation} \\
\textit{Chinese Academy of Sciences}\\
Beijing, China \\
bin.tian@ia.ac.cn}

\and
\IEEEauthorblockN{4\textsuperscript{th} Jun Fu}
\IEEEauthorblockA{\textit{Institute of Automation} \\
\textit{Chinese Academy of Sciences}\\
Beijing, China \\
fujun2023@ia.ac.cn}
}

\maketitle

\begin{abstract}
Autonomous parking remains a critical yet challenging task in intelligent driving systems, particularly within constrained urban environments where maneuvering space is limited and precise control is essential. While recent advances in end-to-end learning have shown great promise, the lack of high-quality, structured datasets tailored for parking scenarios remains a significant bottleneck.
To address this gap, we present \textit{ParkingScenes}, a comprehensive multimodal dataset specifically designed for end-to-end autonomous parking in simulated scenes. Built on the CARLA simulator, \textit{ParkingScenes} features structured parking trajectories generated by a Hybrid A* planner and a Model Predictive Controller (MPC), providing accurate and reproducible supervision signals. The dataset includes 16 reverse-in and 6 parallel parking scenarios, each executed under two pedestrian conditions (present and absent), resulting in 704 structured episodes and approximately 105,000 frames. Each scenario is repeated 16 times to ensure consistent coverage. The data collection pipeline is fully open-sourced, supporting scalable expansion and reproducible research.
Each frame contains synchronized data from four RGB cameras, four depth sensors, vehicle motion states, and Bird’s-Eye View (BEV) representations, enabling rich multimodal fusion and context-aware learning. To demonstrate the utility of our dataset, we compare models trained on \textit{ParkingScenes} with those trained on unstructured, manually collected simulation data under identical conditions. Results show significant improvements in performance, underscoring the effectiveness of structured supervision for robust and accurate parking policy learning.
By releasing both the dataset and the collection framework, \textit{ParkingScenes} establishes a scalable and reproducible benchmark for advancing learning-based autonomous parking systems. The dataset and collection framework will be released at: \url{https://github.com/haonan-ai/ParkingScenes}.
\end{abstract}


\section{Introduction}
Autonomous parking is a fundamental capability for intelligent vehicles, especially in urban parking lots where space is constrained and perception accuracy is critical~\cite{b1}. Traditional automated parking systems rely on modular pipelines that separately handle perception, localization, path planning, and control~\cite{b2}. Although these systems are effective in structured environments, they often lack robustness and adaptability in complex or dynamic parking scenarios due to accumulated error, sensor noise, or rule-based decision boundaries~\cite{b3}.

In recent years, end-to-end learning has emerged as a promising paradigm for autonomous driving tasks~\cite{b4}. By directly mapping sensor inputs to control commands, this approach eliminates the need for hand-crafted planning and control modules, offering potential advantages in scalability and performance~\cite{b5}. Several studies have demonstrated the feasibility of using end-to-end networks for tasks such as lane keeping, obstacle avoidance, and autonomous parking in simulation environments~\cite{b6}.

However, the performance of end-to-end models critically depends on the quality, structure, and diversity of the training data~\cite{b7}. Existing datasets for autonomous driving primarily focus on highway or urban navigation, lacking fine-grained parking maneuvers, synchronized control supervision, and dynamic interaction modeling. Prior efforts, such as E2E Parking~\cite{e2e_parking}, collected simulation data manually to support parking policy learning, but the dataset was not publicly released—hindering reproducibility and standardized evaluation. Moreover, human-driven data collection often suffers from limited scenario coverage, inconsistent or suboptimal trajectories, and under-representation of rare corner cases, which collectively impede the development of robust and generalizable parking models.

To address these challenges, we introduce \textit{ParkingScenes}, a new multimodal dataset specifically tailored for end-to-end autonomous parking. Built entirely within the CARLA simulator~\cite{carla}, \textit{ParkingScenes} leverages a hybrid A* planner~\cite{hybrid_A} and a model predictive controller (MPC)~\cite{mpc} to generate high-quality, repeatable parking maneuvers under diverse conditions. Each episode contains synchronized surround-view RGB images, depth maps, Bird’s-Eye View (BEV) representations, vehicle motion states, and dynamic pedestrian information. Covering 16 reverse-in and 6 parallel parking scenarios, each under both pedestrian-present and pedestrian-free conditions, the dataset comprises 704 structured episodes and approximately 105,000 annotated frames.

\begin{table*}[t]
\caption{Comparison of Simulated Datasets for Autonomous Parking Scenarios}
\begin{center}
\begin{tabular}{|l|c|c|c|c|c|c|c|}
\hline
\textbf{Dataset} & \textbf{Simulation} & \textbf{Parking Env.} & \textbf{Parking Behavior} & \textbf{Control Data} & \textbf{Trajectory Data} & \textbf{Surround-view} & \textbf{Public} \\
\hline
KITTI~\cite{kitti}           & \ding{55} & \ding{55} & \ding{55} & \ding{55} & \ding{51} & \ding{55} & \ding{51} \\
\hline
nuScenes~\cite{nuscenes}     & \ding{55} & \ding{55} & \ding{55} & \ding{51} & \ding{51} & \ding{51} & \ding{51} \\
\hline
Waymo Open~\cite{waymo}           & \ding{55} & \ding{55} & \ding{55} & \ding{55} & \ding{51} & \ding{51} & \ding{51} \\
\hline
BDD100K~\cite{bdd100k}       & \ding{55} & \ding{55} & \ding{55} & \ding{55} & \ding{51} & \ding{55} & \ding{51} \\
\hline
Cityscapes~\cite{cityscapes} & \ding{55} & \ding{55} & \ding{55} & \ding{55} & \ding{55} & \ding{55} & \ding{51} \\
\hline
PSV~\cite{psv_dataset}       & \ding{55} & \ding{51} & \ding{55} & \ding{55} & \ding{55} & \ding{51} & \ding{55} \\
\hline
DeepPS~\cite{deep_ps_dataset}& \ding{55} & \ding{51} & \ding{55} & \ding{55} & \ding{55} & \ding{51} & \ding{51} \\
\hline
SNU Parking~\cite{snu_parking} & \ding{55} & \ding{51} & \ding{55} & \ding{55} & \ding{55} & \ding{51} & \ding{51} \\
\hline
DLP~\cite{dlp_dataset}       & \ding{55} & \ding{51} & \ding{55} & \ding{55} & \ding{51} & \ding{55} & \ding{51} \\
\hline
Audi A2D2~\cite{a2d2}        & \ding{55} & \ding{55} & \ding{55} & \ding{51} & \ding{51} & \ding{51} & \ding{51} \\
\hline
Comma2k19~\cite{comma2k19}   & \ding{55} & \ding{55} & \ding{55} & \ding{51} & \ding{51} & \ding{55} & \ding{51} \\
\hline
SUPS~\cite{sups_dataset}     & \ding{51} & \ding{51} & \ding{55} & \ding{55} & \ding{51} & \ding{51} & \ding{51} \\
\hline
E2E Parking~\cite{e2e_parking} & \ding{51} & \ding{51} & \ding{51} & \ding{51} & \ding{51} & \ding{51} & \ding{55} \\
\hline
\textbf{ParkingScenes (Ours)}       & \ding{51} & \ding{51} & \ding{51} & \ding{51} & \ding{51} & \ding{51} & \ding{51} \\
\hline
\multicolumn{8}{|l|}{\ding{51}: Available, \ding{55}: Not Available.} \\
\hline
\end{tabular}
\label{tab:dataset_comparison}
\end{center}
\end{table*}

\begin{figure}[t]
\centerline{\includegraphics[width=0.48\textwidth]{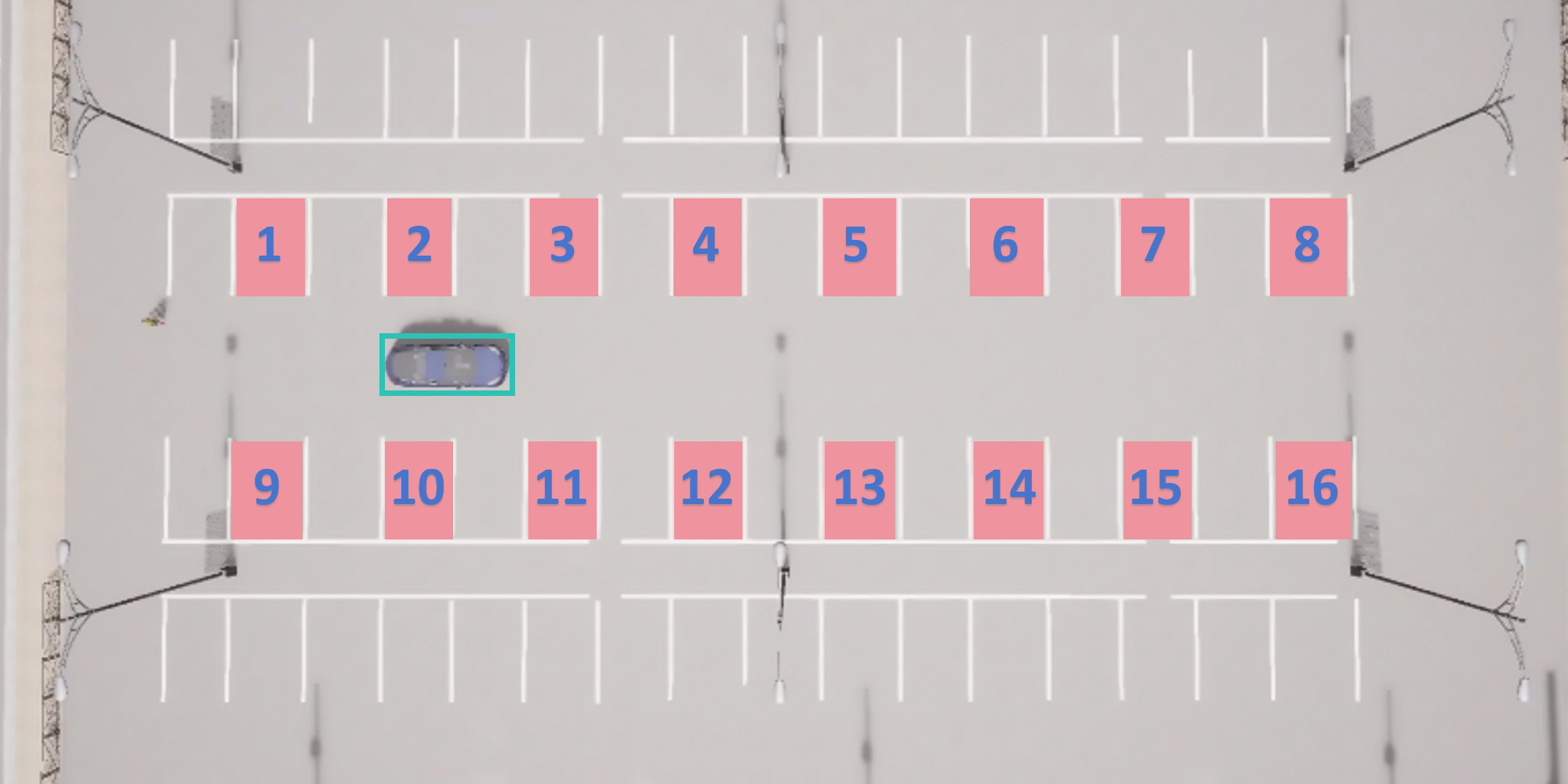}}
\caption{Top-down view of 16 reverse-in parking slots in simulation. Parking slots are visualized in pink, and the ego vehicle is marked in green.}
\label{fig:reverse_parking}
\end{figure}

We train an end-to-end neural network using the \textit{ParkingScenes} dataset and compare its performance against models trained on unstructured, manually collected simulation data under identical conditions. Experimental results demonstrate that, despite using the same architecture and training procedure, models trained on \textit{ParkingScenes} notably improve success rate, precision, and robustness. These results highlight the benefits of structured, automated trajectory generation and confirm the dataset’s value for learning-based autonomous parking research.

Our main contributions are summarized as follows:

\begin{itemize}
    \item We present \textit{ParkingScenes}, a comprehensive multimodal dataset for end-to-end autonomous parking in simulation, providing high-resolution surround-view imagery, depth, vehicle states, and BEV representations across diverse parking scenarios. The dataset is publicly available to facilitate benchmarking and reproducible research.

    \item We develop a structured data collection pipeline that integrates a hybrid A* planner and Model Predictive Control (MPC), enabling the generation of high-quality, repeatable trajectories. While both methods are well-established, their integration into a unified and fully automated data generation framework for autonomous parking is, to the best of our knowledge, the first of its kind. The entire pipeline is open-sourced to support scalable dataset expansion and controlled experimentation.

    \item We demonstrate that models trained on \textit{ParkingScenes} significantly outperform those trained on manually collected simulation data under identical conditions, validating the effectiveness of structured supervision for end-to-end autonomous parking. This highlights our key contribution: showing that the quality and consistency of training data play a decisive role in improving the reliability of end-to-end parking policies.
\end{itemize}

\begin{figure}[t]
\centerline{\includegraphics[width=0.48\textwidth]{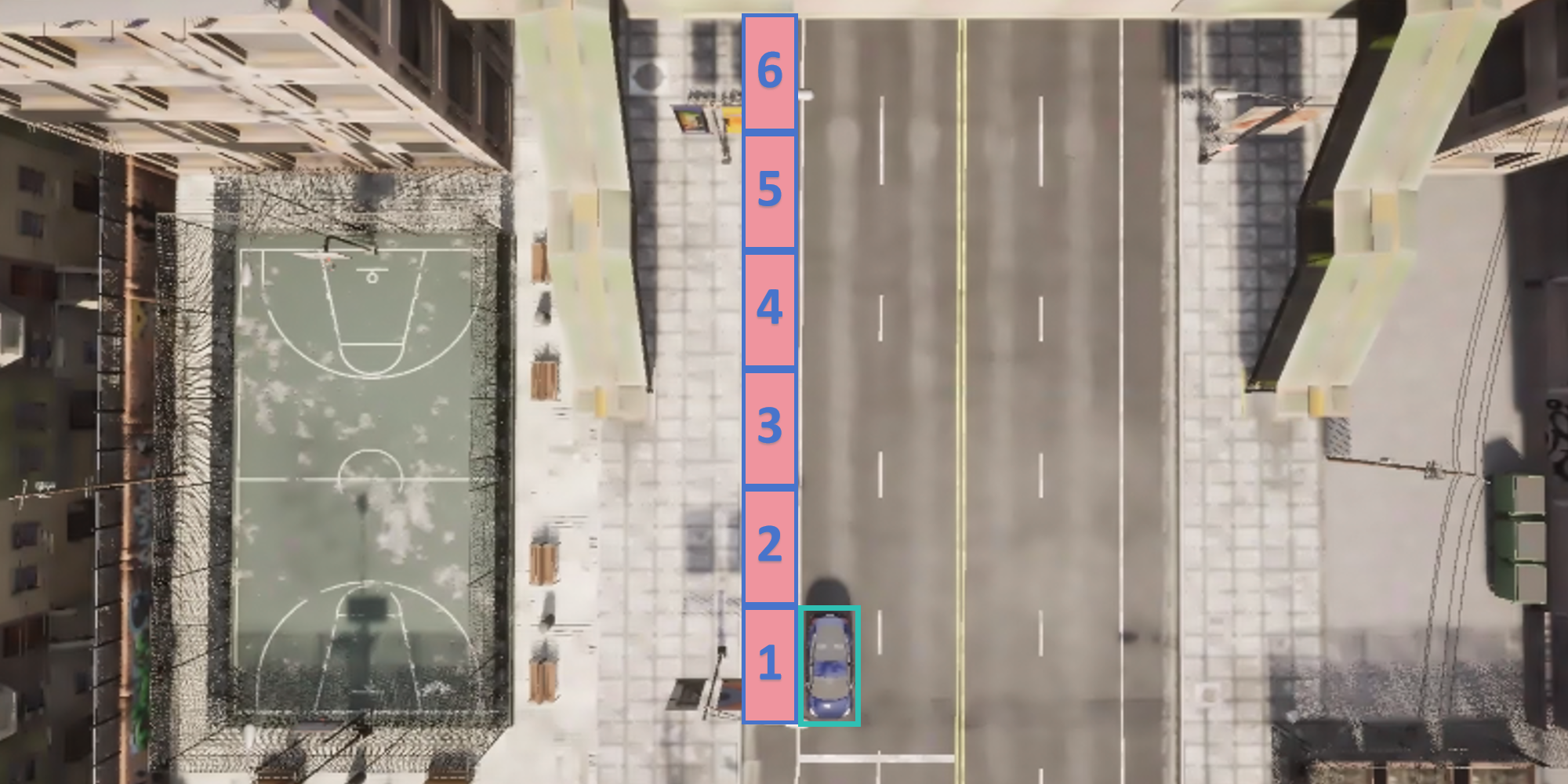}}
\caption{Top-down view of 6 parallel parking slots in simulation. The layout reflects a typical roadside parking scenario with marked slots and vehicle entry points.}
\label{fig:parallel_parking}
\end{figure}

\begin{figure}[t]
\centering
\subfloat[Top-down: No pedestrian]{%
  \includegraphics[width=0.48\columnwidth]{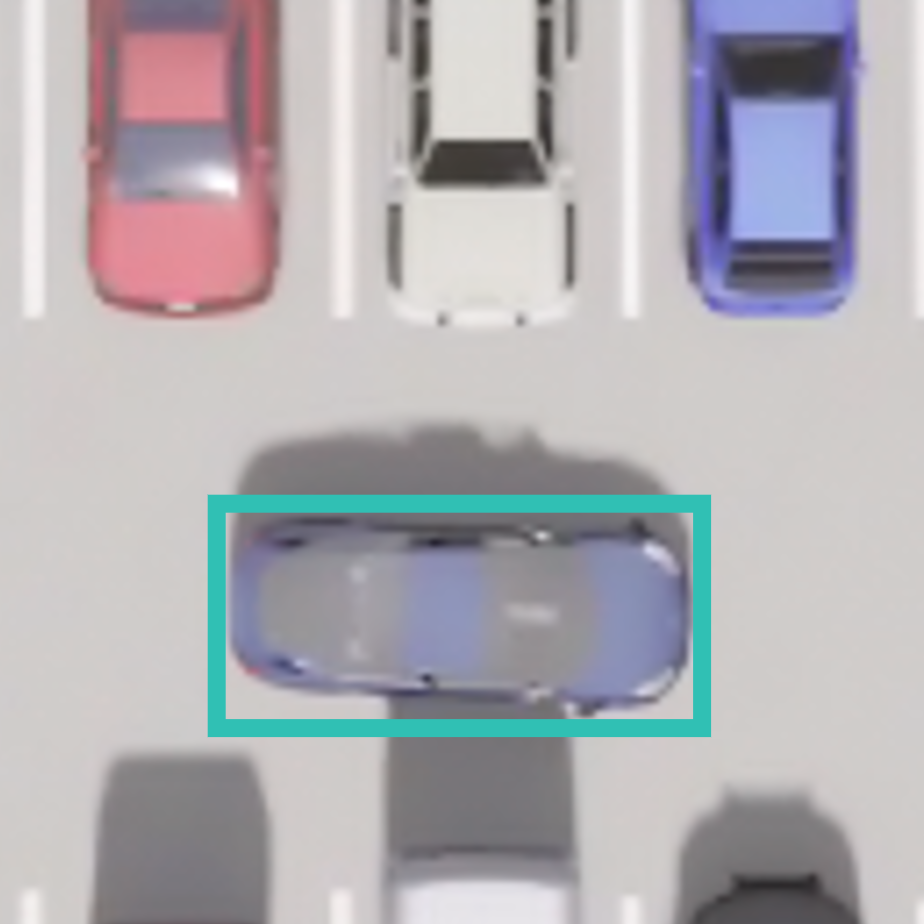}
}
\hfill
\subfloat[Top-down: With pedestrian]{%
  \includegraphics[width=0.48\columnwidth]{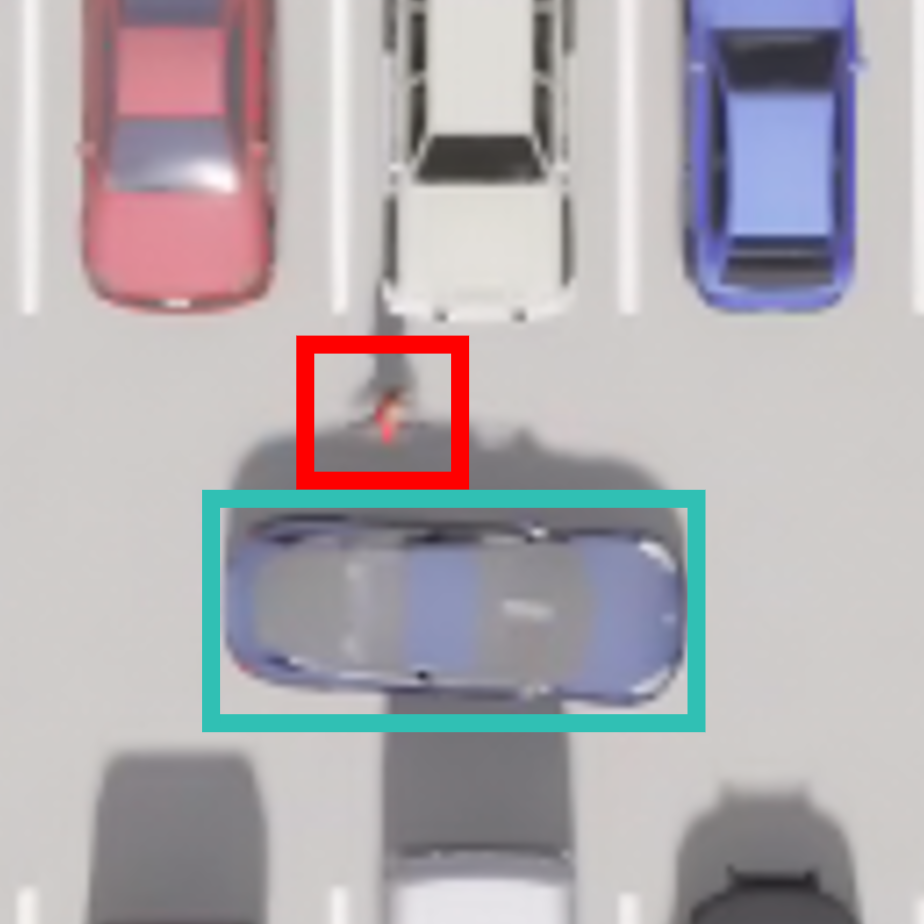}
}
\vspace{5pt}
\subfloat[Rear-view: No pedestrian]{%
  \includegraphics[width=0.48\columnwidth]{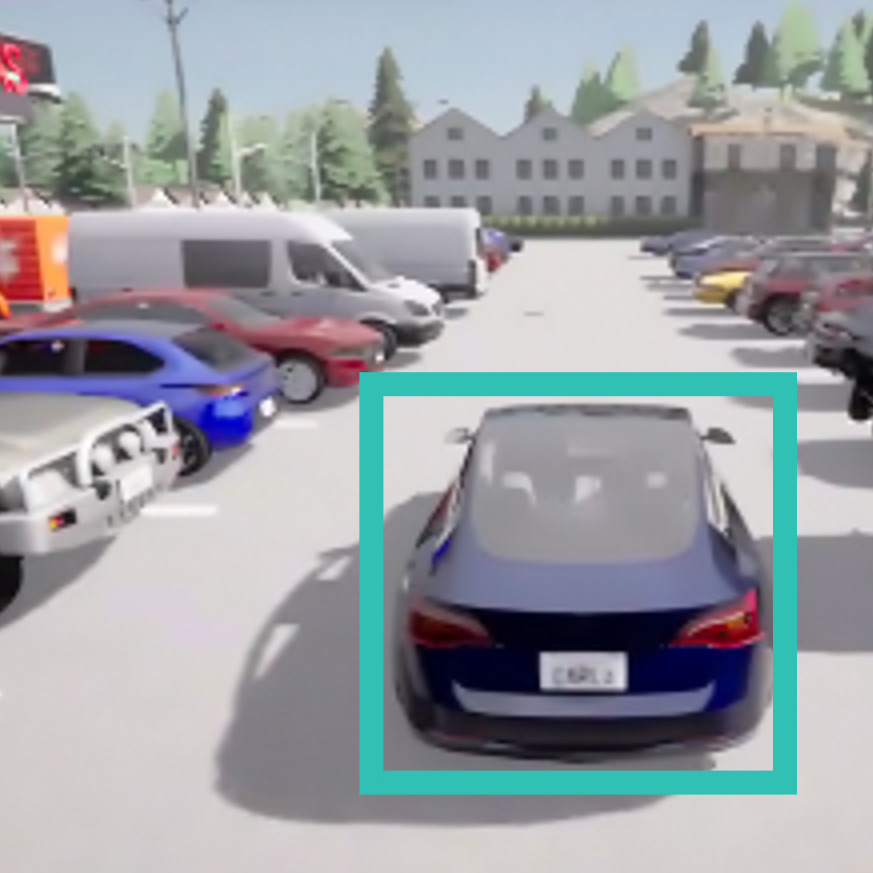}
}
\hfill
\subfloat[Rear-view: With pedestrian]{%
  \includegraphics[width=0.48\columnwidth]{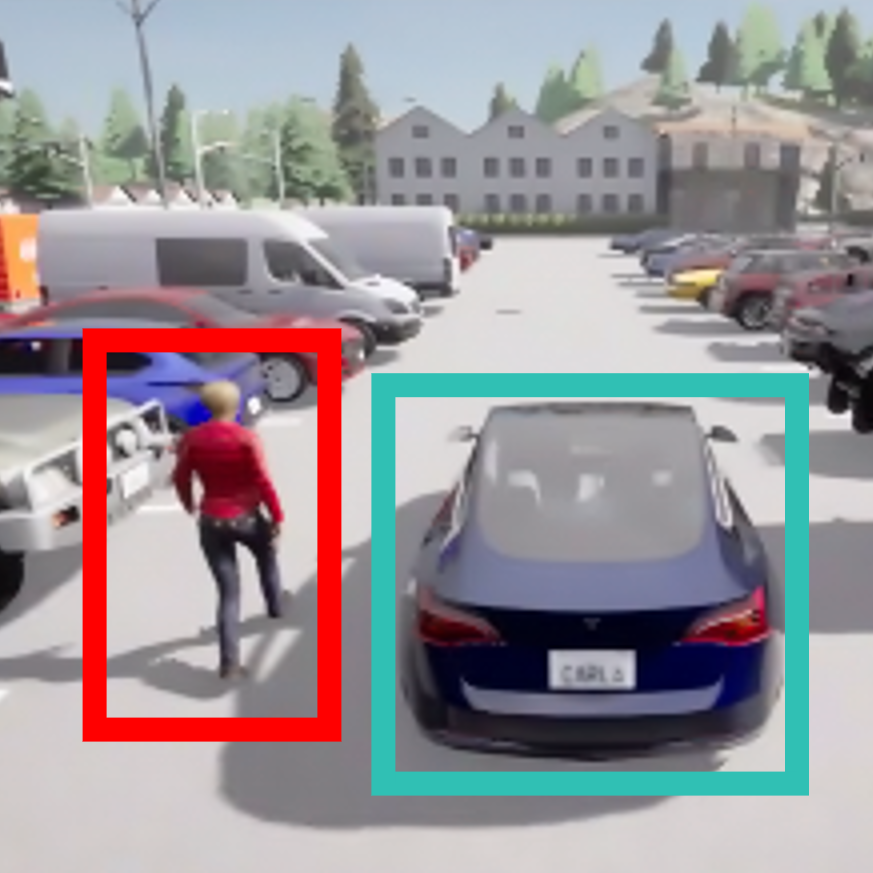}
}
\caption{Comparison of parking scenarios with and without pedestrians, captured from top-down and rear-view perspectives. The ego vehicle is outlined in green, while dynamic pedestrians are marked with red bounding boxes.}
\label{fig:pedestrians}
\end{figure}

\section{Related Work}

A comparison of existing autonomous driving datasets related to parking is shown in Table~\ref{tab:dataset_comparison}. Many well-known datasets primarily target urban navigation or general driving perception, but lack structured parking maneuvers, low-speed control data, or full surround-view sensing.

Large-scale benchmarks such as KITTI~\cite{kitti}, nuScenes~\cite{nuscenes}, Waymo Open~\cite{waymo}, BDD100K~\cite{bdd100k}, and Cityscapes~\cite{cityscapes} have greatly advanced urban driving research, yet none include parking-specific behaviors or control-level supervision.

Some datasets specifically address parking perception. PSV~\cite{psv_dataset}, DeepPS~\cite{deep_ps_dataset}, and SNU Parking~\cite{snu_parking} provide surround-view imagery and slot annotations, but lack temporal vehicle states and control signals. DLP~\cite{dlp_dataset} offers trajectory annotations from drone videos, while Audi A2D2~\cite{a2d2} and Comma2k19~\cite{comma2k19} include control data but are not captured in parking environments.

Simulation-based datasets begin to fill this gap. SUPS~\cite{sups_dataset} provides synthetic underground environments, while E2E Parking~\cite{e2e_parking} uses CARLA for manually collected parking trajectories. However, these efforts either lack control signals or are not publicly released. 

In contrast, \textit{ParkingScenes} is the first to combine surround-view multimodal sensing, structured and repeatable maneuvers, synchronized control supervision, and full public accessibility within a unified simulated platform, establishing a comprehensive benchmark for autonomous parking research.

\begin{figure}[t]
\centering
\subfloat[Front-view RGB image]{%
  \includegraphics[width=0.48\linewidth]{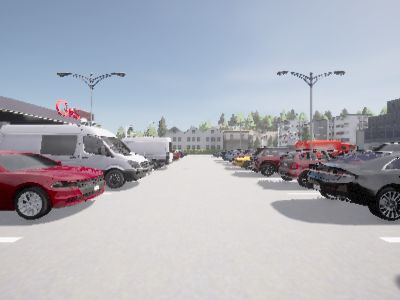}
}
\hfill
\subfloat[Depth map]{%
  \includegraphics[width=0.48\linewidth]{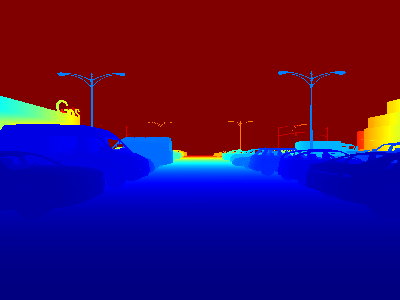}
}
\vspace{4pt}
\subfloat[Top-down RGB (BEV camera)]{%
  \includegraphics[width=0.48\linewidth]{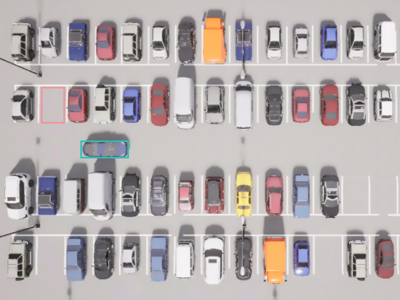}
}
\hfill
\subfloat[Semantic BEV representation]{%
  \includegraphics[width=0.48\linewidth]{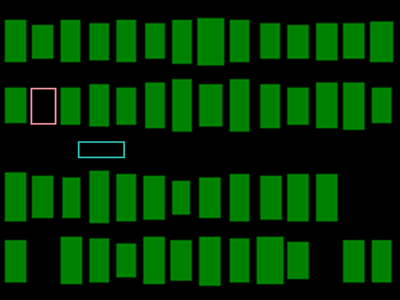}
}
\caption{
Multimodal views captured per frame in our dataset. Top row: front-view RGB image (a) and corresponding depth map (b). Bottom row: a top-down RGB view rendered from CARLA’s Spectator camera (c), and a semantic BEV map generated by our custom renderer (d). Green box outlines denote the ego vehicle, pink boxes mark the target parking slot, and filled green rectangles indicate static vehicles; all are manually added for illustration.
}
\label{fig:modalities_four}
\end{figure}

\section{Simulation Dataset}

To support learning-based autonomous parking, we construct a structured and scalable dataset entirely within the CARLA simulator~\cite{carla}. The dataset captures diverse parking maneuvers under controlled and reproducible conditions, ensuring high-quality supervision for training and benchmarking end-to-end models.

The spatial configurations of the simulated parking scenarios are illustrated in Fig.~\ref{fig:reverse_parking} and Fig.~\ref{fig:parallel_parking}. We simulate two primary parking types, namely reverse-in and parallel parking, each instantiated across multiple slot layouts embedded in urban-like environments to mimic realistic conditions. Specifically, the reverse-in parking setup covers 16 distinct slots, while the parallel parking setup includes 6 distinct slots.

To introduce dynamic complexity and enhance behavioral diversity, we design environmental variations featuring the presence and absence of pedestrian traffic. Representative examples are illustrated in Fig.~\ref{fig:pedestrians}. For each condition (with or without pedestrians), every parking scenario is executed 16 times from randomized initial states, resulting in 22 distinct scenario types and a total of 704 structured parking episodes, encompassing approximately 105,000 frames. 
The randomized initial states are sampled within a bounded range of entry positions and orientations relative to the target slot, which introduces trajectory diversity while ensuring feasibility and comparability across episodes.

In each simulation frame, we capture multimodal sensor observations, including RGB images, depth images, and top-down spatial representations, as shown in Fig.~\ref{fig:modalities_four}. These modalities provide complementary perspectives for perception, scene understanding, and trajectory supervision.

\begin{figure}[t]
\centering
\includegraphics[width=0.48\textwidth]{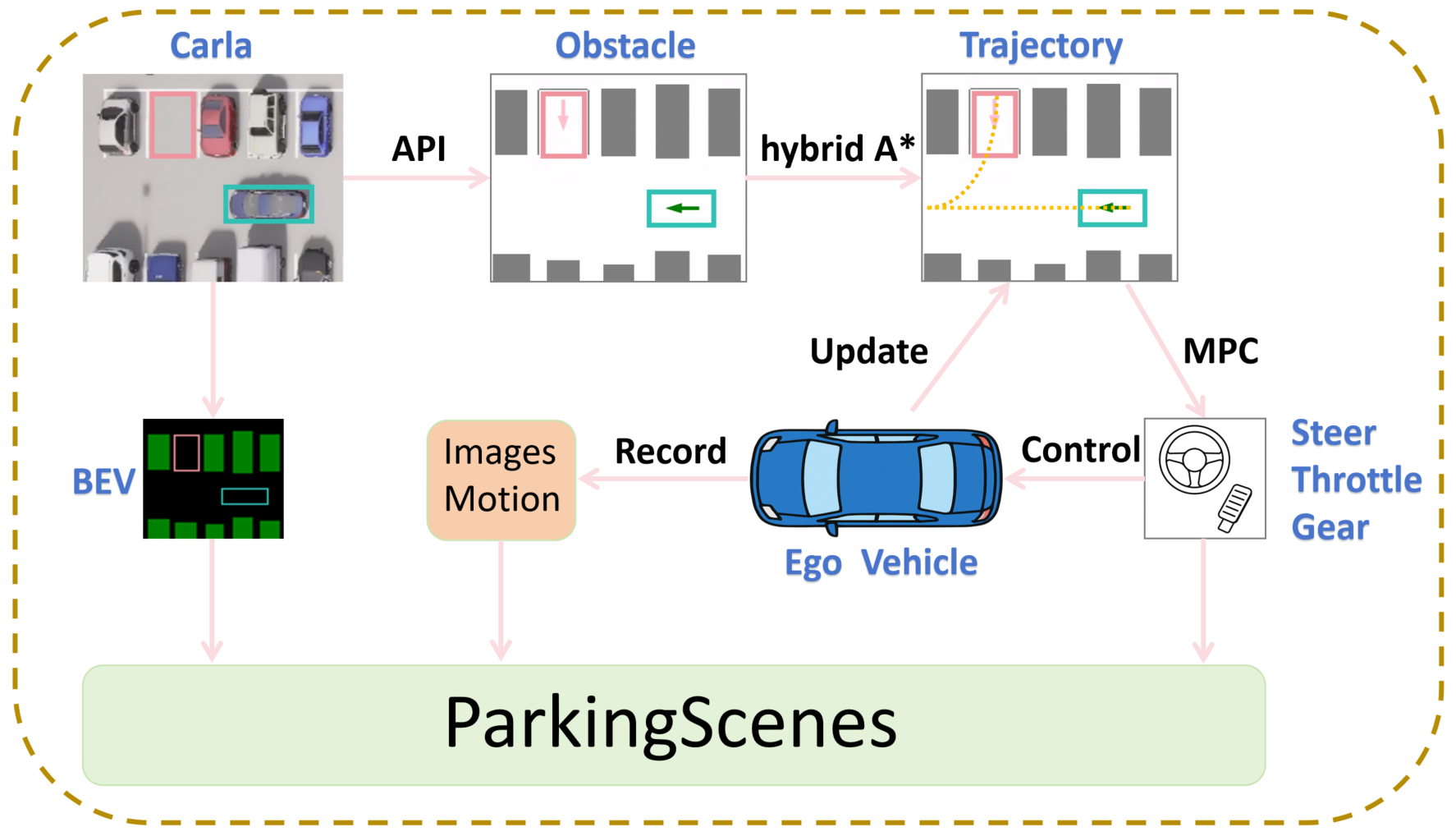}
\caption{Overview of the ParkingScenes data collection pipeline. The ego vehicle, obstacles, and parking slot are initialized in CARLA. Scene information is processed by a Hybrid A* planner and tracked by an MPC controller in a closed-loop fashion. Multimodal data are synchronized and recorded at each step.}
\label{fig:collection_flow}
\end{figure}

Each frame contains the following synchronized data:

\begin{itemize}
    \item Four surround-view RGB images capturing 360° visual context;
    \item Four depth images providing geometric structure;
    \item Bird’s-Eye View (BEV) representations generated from semantic and spatial projections;
    \item Vehicle motion states, including position, speed, acceleration, and steering angle.
\end{itemize}

Since all data are generated within the CARLA simulator, sensor measurements do not suffer from real-world noise or distance-dependent accuracy attenuation. However, depth values inherently encode distance information, which can be leveraged to study distance-related perception effects. In future extensions to real-world data, physical sensor attenuation characteristics at different ranges will be explicitly considered.

All data are collected under fixed environmental settings, including consistent daylight illumination, to ensure comparability across episodes. Although lighting variations such as dusk or night can be configured in CARLA, we intentionally keep lighting fixed in this release to focus on structured trajectory supervision. Ground-truth control signals—including throttle, brake, and steering—are directly aligned with the visual and spatial context through automated execution. This setup supports sensor fusion tasks and facilitates robust perception-to-control learning. By combining scalable simulation with a structured control architecture, the dataset provides a clean benchmark for evaluating end-to-end autonomous parking systems across varied yet reproducible scenarios.

\section{Collection Pipeline}

We provide an overview of the automated data collection and control pipeline, as illustrated in Fig.~\ref{fig:collection_flow}. The pipeline operates entirely within the CARLA simulator and follows a structured process. It begins with the initialization of the ego vehicle, surrounding obstacles, and the target parking slot. Scene information is retrieved via the CARLA API and passed to a Hybrid A* planner~\cite{hybrid_A}, which generates a collision-free global path. A Model Predictive Controller (MPC)~\cite{mpc} then tracks the reference trajectory and issues real-time control commands. This closed-loop control cycle repeats at each simulation step, producing structured and consistent parking behaviors. Throughout execution, synchronized multimodal data—including bird's-eye view (BEV) maps, RGB and depth images, vehicle states, and control signals—are continuously recorded to form the ParkingScenes dataset.

\begin{figure}[t]
\centering
\includegraphics[width=0.48\textwidth]{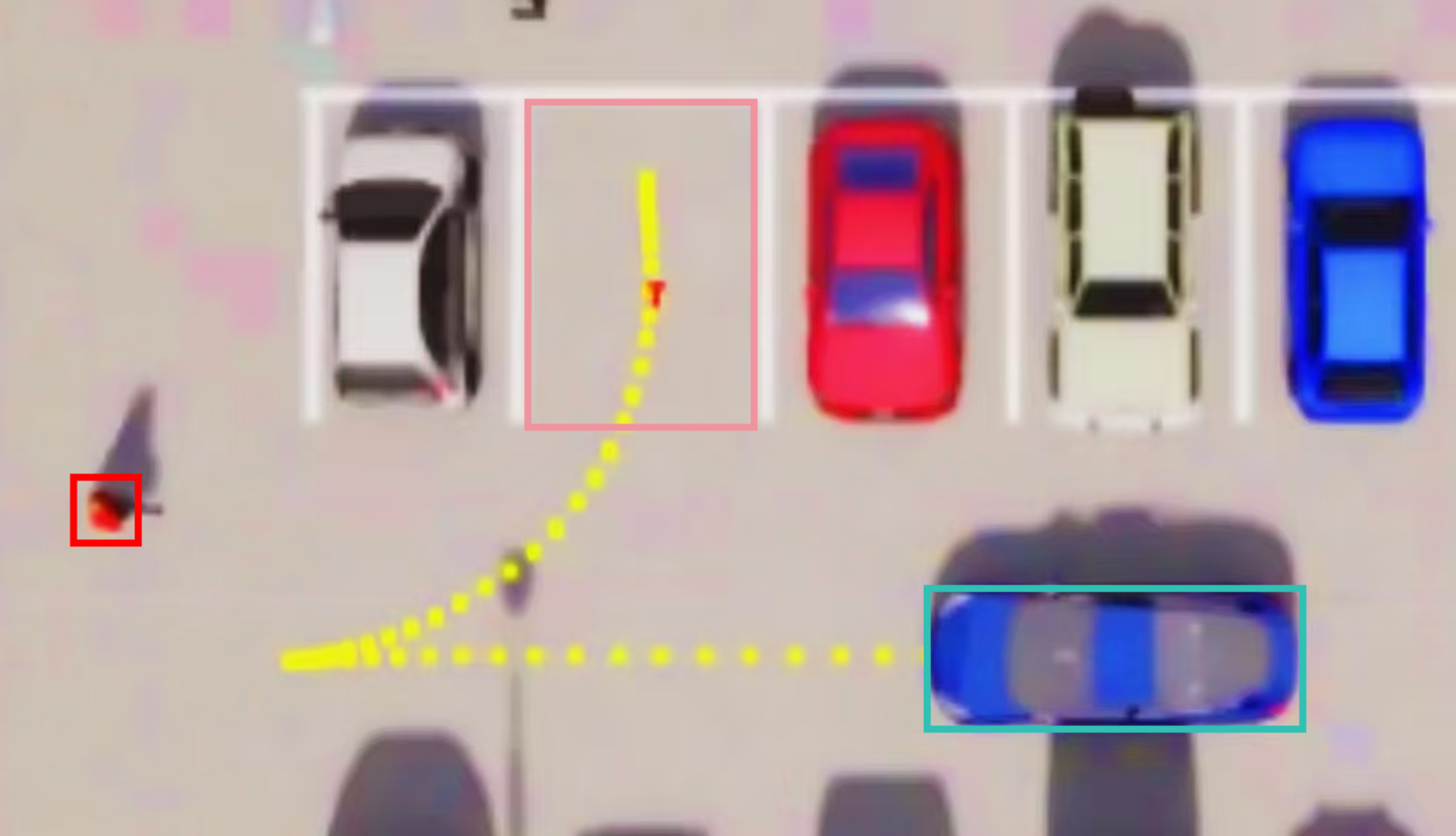}
\caption{Top-down visualization of a planned reverse-in parking trajectory in the CARLA environment. The ego vehicle is outlined in green, the target parking slot in pink, pedestrians in red, and the trajectory as yellow dots.}
\label{fig:planned_trajectory}
\end{figure}

An example reverse-in parking scenario generated by the pipeline is shown in Fig.~\ref{fig:planned_trajectory}, where the ego vehicle is outlined in green, the target parking slot in pink, dynamic pedestrians in red, and the planned trajectory is visualized as yellow dots. This case demonstrates the structured and repeatable motion patterns achieved under realistic simulation conditions.

To enable scalable and consistent trajectory generation, we implement a fully automated pipeline that combines a global Hybrid A* planner with a local Model Predictive Controller (MPC), both adapted from prior work. The Hybrid A* module generates kinematically feasible, obstacle-avoiding paths between the initial and goal poses. The MPC then tracks the reference trajectories by optimizing steering and throttle inputs over a short prediction horizon, ensuring smooth and precise control. In practice, the interaction between the two modules is stabilized by co-tuning a small set of parameters: on the MPC side, the prediction horizon and sampling time (\(N, \Delta t\)), state/control weights (e.g., \(Q, R\)), steering-rate penalties, and input/speed saturations; on the Hybrid A* side, the grid and heading resolutions, curvature/reverse/switch-back penalties, and goal tolerances. We use fixed defaults (as released with our configs) and adjust them only according to the curvature and velocity profile of the planned path, which balances global feasibility with local tracking accuracy.

Furthermore, our pipeline supports flexible scenario configuration, including:
\begin{itemize}
    \item Parking slot layouts with varied orientations and dimensions;
    \item Dynamic pedestrian injection for behavioral diversity;
    \item Customizable sensor setups for surround-view and depth imaging;
    \item Randomized initial and goal poses to enhance generalization.
\end{itemize}

All control signals, vehicle states, and multimodal sensor data are automatically logged and temporally synchronized at each frame. In total, we collect 704 structured episodes across 22 distinct simulation settings, resulting in over 105,000 frames. While the current release focuses on reverse-in and parallel parking scenarios as fundamental behaviors, the pipeline design supports extension to more complex layouts, such as angled or multi-vehicle parking, which we plan to explore in future work.

We release both the dataset and the collection pipeline to promote reproducibility and extensibility within the research community.

\section{Experiments and Results}

\subsection{Experimental Setup}

To evaluate the effectiveness of structured parking trajectories in training end-to-end autonomous parking models, we adopt the E2E Parking framework~\cite{e2e_parking} as our baseline architecture. The model takes four surround-view RGB images and basic vehicle motion states as input, and outputs discrete control signals including acceleration, steering angle, and gear selection. We retain the original network design, including BEV generation via LSS, Transformer-based feature fusion and control decoding, and auxiliary depth prediction.

All experiments are conducted using the CARLA 0.9.11 simulator. During training, the model is supervised using our structured reverse-in parking trajectories collected under pedestrian-free conditions via a Hybrid A* planner and a Model Predictive Controller (MPC). To ensure a fair comparison, we constrain our dataset to parking scenes and layout configurations similar to those described in E2E Parking~\cite{e2e_parking}. 

We compare the performance of models trained on our dataset against the reported results from E2E Parking, where the model was trained using manually collected simulation trajectories. The training settings, including optimizer parameters, image resolution, and batch size, follow the original implementation~\cite{e2e_parking}. Closed-loop evaluation is conducted across 16 distinct reverse-in parking scenes, each with 8 repetitions from randomized starting positions.

\subsection{Baseline: Performance on Human-Collected Data}

We first summarize the reported performance of the baseline model trained on manually collected trajectories from the original E2E Parking paper~\cite{e2e_parking}. In their closed-loop experiments, the model achieved an average Target Success Rate (TSR) of 91.41\%, with an Average Position Error (APE) of 0.30 meters and an Average Orientation Error (AOE) of 0.87 degrees. The Collision Rate (CR) was 2.08\%, and the Non-Target Rate (NTR) was 3.39\%, reflecting occasional failures caused by trajectory inconsistency or suboptimal maneuvers. According to their report, the training data consisted of approximately 22,000 frames collected manually by four experienced drivers.

\subsection{Our Results: Performance on Structured Data}

We then train the same model architecture on our structured dataset, which consists of automated reverse-in parking trajectories under identical conditions but without pedestrians. The trajectory generation is executed by a Hybrid A* global planner and a Model Predictive Controller (MPC), yielding smooth, collision-free paths with minimal noise and variability. This setting serves as a clean benchmark for supervised learning.

After training, we perform the same closed-loop evaluation protocol as the baseline. Our model achieves a Target Success Rate of \textbf{96.35\%}, significantly improving upon the baseline. The Average Orientation Error (AOE) is reduced to \textbf{0.61 degrees}, and the Average Position Error (APE) decreases to \textbf{0.24 meters}. The Collision Rate drops to \textbf{1.04\%}, indicating enhanced safety and stability during execution. These results demonstrate that training on structured, noise-free data can yield more reliable and accurate parking behaviors, even when evaluated in randomized test scenarios.

\begin{table}[t]
\caption{Closed-Loop Evaluation: Structured vs. Human-Collected Data}
\centering
\begin{tabular}{|l|c|c|}
\hline
\textbf{Metric} & \textbf{E2E Parking} & \textbf{ParkingScenes} \\
\hline
Target Success Rate (TSR) ↑ & 91.41\% & \textbf{96.35\%} \\
\hline
Target Failure Rate (TFR) ↓ & 2.08\% & \textbf{1.04\%} \\
\hline
Non-Target Rate (NTR) ↓     & 3.39\% & \textbf{1.56\%} \\
\hline
Collision Rate (CR) ↓       & 2.08\% & \textbf{1.04\%} \\
\hline
Timeout Rate (TR) ↓         & 1.04\% & \textbf{0.52\%} \\
\hline
Avg. Position Error (APE) ↓ & 0.30 m & \textbf{0.24 m} \\
\hline
Avg. Orientation Error (AOE) ↓ & 0.87$^\circ$ & \textbf{0.61$^\circ$} \\
\hline
Avg. Parking Time (APT) ↓   & 15.72 s & \textbf{14.98 s} \\
\hline
\end{tabular}
\label{tab:structured_vs_human}
\end{table}

\subsection{Comparative Analysis: Impact of Structured Trajectories}

The improvement observed across all performance metrics highlights the advantage of using structured trajectories for training. Unlike human-collected data, which may contain minor oscillations, late corrections, or inconsistent control signals, the trajectories generated by our Hybrid A* + MPC pipeline are smooth, reproducible, and optimized for parking precision.

These consistent and low-noise supervision signals facilitate stable gradient updates, leading to faster convergence and improved generalization. Moreover, because our data is free from control drift or visual misalignment, the model learns a more accurate mapping between visual cues (e.g., parking slot lines, lane boundaries) and control outputs. As a result, the network exhibits improved consistency and a higher likelihood of reaching the intended target slot within defined spatial and temporal tolerances.

\subsection{Discussion}

While structured data clearly benefits training stability and execution accuracy, it also reveals potential limitations. Specifically, the current experiments are restricted to reverse-in parking scenarios without pedestrians. Future work will extend this evaluation to more complex environments, including dynamic obstacles and occluded parking slots. We also plan to investigate mixed training strategies that combine structured and manually collected data to assess robustness under distributional shifts and behavioral variability.

In summary, these results demonstrate that data quality and trajectory consistency play a critical role in end-to-end parking systems. Structured parking trajectories not only yield better quantitative performance, but also support more reproducible and scalable training pipelines. This provides new evidence for the broader autonomous driving community that improving dataset structure and supervision quality can directly translate into more reliable learning outcomes.

\section{Conclusion}

In this paper, we introduced \textit{ParkingScenes}, a comprehensive multimodal dataset tailored for end-to-end autonomous parking in simulation. By leveraging structured trajectory generation via Hybrid A* and MPC, our dataset enables scalable, high-fidelity, and reproducible training for data-driven parking systems. Extensive experiments demonstrate that models trained on structured data not only outperform those trained on manually collected trajectories, but also exhibit greater robustness and precision in closed-loop evaluations.

Compared to existing autonomous driving datasets, \textit{ParkingScenes} uniquely focuses on parking-specific behaviors, full surround-view imagery, and synchronized control annotations, all within a fully simulated and controllable environment. Through careful design of scenes, sensor configurations, and trajectory consistency, we provide a high-quality benchmark for evaluating perception, control, and sensor fusion in the context of autonomous parking.

Looking forward, we plan to expand \textit{ParkingScenes} in several directions. First, we aim to introduce more dynamic simulation environments, including moving pedestrians and interactive vehicles. Second, although the current validation is performed in simulation, we will explore the collection of real-world parking data to support sim-to-real transfer learning. Finally, we intend to release baseline models and evaluation tools to promote community adoption and facilitate fair comparisons across different learning paradigms.

By releasing \textit{ParkingScenes}, we hope to foster future research in scalable, reliable, and transferable autonomous parking systems, and contribute to the broader development of intelligent vehicle autonomy in constrained urban environments.

\bibliographystyle{IEEEtran}
\bibliography{references}

@inproceedings{b1,
  title={Avp-slam: Semantic visual mapping and localization for autonomous vehicles in the parking lot},
  author={Qin, Tong and Chen, Tongqing and Chen, Yilun and Su, Qing},
  booktitle={2020 IEEE/RSJ International Conference on intelligent robots and systems (IROS)},
  pages={5939--5945},
  year={2020},
  organization={IEEE}
}

@article{b2,
  title={Robust parking path planning with error-adaptive sampling under perception uncertainty},
  author={Lee, Seongjin and Lim, Wonteak and Sunwoo, Myoungho},
  journal={Sensors},
  volume={20},
  number={12},
  pages={3560},
  year={2020},
  publisher={MDPI}
}

@article{b3,
  title={Model-Based predictive control and reinforcement learning for planning vehicle-parking trajectories for vertical parking spaces},
  author={Shi, Junren and Li, Kexin and Piao, Changhao and Gao, Jun and Chen, Lizhi},
  journal={Sensors},
  volume={23},
  number={16},
  pages={7124},
  year={2023},
  publisher={MDPI}
}

@article{b4,
  title={Recent advancements in end-to-end autonomous driving using deep learning: A survey},
  author={Chib, Pranav Singh and Singh, Pravendra},
  journal={IEEE Transactions on Intelligent Vehicles},
  volume={9},
  number={1},
  pages={103--118},
  year={2023},
  publisher={IEEE}
}

@inproceedings{b5,
  title={Planning-oriented autonomous driving},
  author={Hu, Yihan and Yang, Jiazhi and Chen, Li and Li, Keyu and Sima, Chonghao and Zhu, Xizhou and Chai, Siqi and Du, Senyao and Lin, Tianwei and Wang, Wenhai and others},
  booktitle={Proceedings of the IEEE/CVF conference on computer vision and pattern recognition},
  pages={17853--17862},
  year={2023}
}

@article{b6,
  title={Virtual fluid-flow-model-based lane-keeping integrated with collision avoidance control system design for autonomous vehicles},
  author={Cheng, Shuo and Li, Liang and Liu, Yong-Gang and Li, Wei-Bing and Guo, Hong-Qiang},
  journal={IEEE Transactions on Intelligent Transportation Systems},
  volume={22},
  number={10},
  pages={6232--6241},
  year={2020},
  publisher={IEEE}
}

@article{b7,
  title={A comprehensive review on Deep Learning-based motion planning and end-to-end learning for self-driving vehicle},
  author={Ganesan, Manikandan and Kandhasamy, Sivanathan and Chokkalingam, Bharatiraja and Mihet-Popa, Lucian},
  journal={IEEE Access},
  year={2024},
  publisher={IEEE}
}

@inproceedings{e2e_parking,
  title={E2e parking: Autonomous parking by the end-to-end neural network on the carla simulator},
  author={Yang, Yunfan and Chen, Denglong and Qin, Tong and Mu, Xiangru and Xu, Chunjing and Yang, Ming},
  booktitle={2024 IEEE Intelligent Vehicles Symposium (IV)},
  pages={2375--2382},
  year={2024},
  organization={IEEE}
}

@inproceedings{kitti,
  title={Are we ready for autonomous driving? the kitti vision benchmark suite},
  author={Geiger, Andreas and Lenz, Philip and Urtasun, Raquel},
  booktitle={2012 IEEE conference on computer vision and pattern recognition},
  pages={3354--3361},
  year={2012},
  organization={IEEE}
}

@inproceedings{nuscenes,
  title={nuscenes: A multimodal dataset for autonomous driving},
  author={Caesar, Holger and Bankiti, Varun and Lang, Alex H and Vora, Sourabh and Liong, Venice Erin and Xu, Qiang and Krishnan, Anush and Pan, Yu and Baldan, Giancarlo and Beijbom, Oscar},
  booktitle={Proceedings of the IEEE/CVF conference on computer vision and pattern recognition},
  pages={11621--11631},
  year={2020}
}

@inproceedings{waymo,
  title={Scalability in perception for autonomous driving: Waymo open dataset},
  author={Sun, Pei and Kretzschmar, Henrik and Dotiwalla, Xerxes and Chouard, Aurelien and Patnaik, Vijaysai and Tsui, Paul and Guo, James and Zhou, Yin and Chai, Yuning and Caine, Benjamin and others},
  booktitle={Proceedings of the IEEE/CVF conference on computer vision and pattern recognition},
  pages={2446--2454},
  year={2020}
}

@inproceedings{cityscapes,
  title={The cityscapes dataset for semantic urban scene understanding},
  author={Cordts, Marius and Omran, Mohamed and Ramos, Sebastian and Rehfeld, Timo and Enzweiler, Markus and Benenson, Rodrigo and Franke, Uwe and Roth, Stefan and Schiele, Bernt},
  booktitle={Proceedings of the IEEE conference on computer vision and pattern recognition},
  pages={3213--3223},
  year={2016}
}

@article{bdd100k,
  title={Bdd100k: A diverse driving video database with scalable annotation tooling},
  author={Yu, Fisher and Xian, Wenqi and Chen, Yingying and Liu, Fangchen and Liao, Mike and Madhavan, Vashisht and Darrell, Trevor and others},
  journal={arXiv preprint arXiv:1805.04687},
  volume={2},
  number={5},
  pages={6},
  year={2018}
}

@inproceedings{carla,
  title={CARLA: An open urban driving simulator},
  author={Dosovitskiy, Alexey and Ros, German and Codevilla, Felipe and Lopez, Antonio and Koltun, Vladlen},
  booktitle={Conference on robot learning},
  pages={1--16},
  year={2017},
  organization={PMLR}
}

@article{hybrid_A,
  title={Practical search techniques in path planning for autonomous driving},
  author={Dolgov, Dmitri and Thrun, Sebastian and Montemerlo, Michael and Diebel, James},
  journal={ann arbor},
  volume={1001},
  number={48105},
  pages={18--80},
  year={2008}
}

@article{mpc,
  title={Constrained model predictive control: Stability and optimality},
  author={Mayne, David Q and Rawlings, James B and Rao, Christopher V and Scokaert, Pierre OM},
  journal={Automatica},
  volume={36},
  number={6},
  pages={789--814},
  year={2000},
  publisher={Elsevier}
}

@article{a2d2,
  title={A2d2: Audi autonomous driving dataset},
  author={Geyer, Jakob and Kassahun, Yohannes and Mahmudi, Mentar and Ricou, Xavier and Durgesh, Rupesh and Chung, Andrew S and Hauswald, Lorenz and Pham, Viet Hoang and M{\"u}hlegg, Maximilian and Dorn, Sebastian and others},
  journal={arXiv preprint arXiv:2004.06320},
  year={2020}
}

@article{comma2k19,
  title={A commute in data: The comma2k19 dataset},
  author={Schafer, Harald and Santana, Eder and Haden, Andrew and Biasini, Riccardo},
  journal={arXiv preprint arXiv:1812.05752},
  year={2018}
}

@inproceedings{psv_dataset,
  title={VH-HFCN based parking slot and lane markings segmentation on panoramic surround view},
  author={Wu, Yan and Yang, Tao and Zhao, Junqiao and Guan, Linting and Jiang, Wei},
  booktitle={2018 IEEE Intelligent Vehicles Symposium (IV)},
  pages={1767--1772},
  year={2018},
  organization={IEEE}
}

@article{deep_ps_dataset,
  title={Vision-based parking-slot detection: A DCNN-based approach and a large-scale benchmark dataset},
  author={Zhang, Lin and Huang, Junhao and Li, Xiyuan and Xiong, Lu},
  journal={IEEE Transactions on Image Processing},
  volume={27},
  number={11},
  pages={5350--5364},
  year={2018},
  publisher={IEEE}
}

@article{snu_parking,
  title={Context-based parking slot detection with a realistic dataset},
  author={Do, Hoseok and Choi, Jin Young},
  journal={IEEE access},
  volume={8},
  pages={171551--171559},
  year={2020},
  publisher={IEEE}
}

@inproceedings{dlp_dataset,
  title={Parkpredict+: Multimodal intent and motion prediction for vehicles in parking lots with cnn and transformer},
  author={Shen, Xu and Lacayo, Matthew and Guggilla, Nidhir and Borrelli, Francesco},
  booktitle={2022 IEEE 25th International Conference on Intelligent Transportation Systems (ITSC)},
  pages={3999--4004},
  year={2022},
  organization={IEEE}
}

@inproceedings{sups_dataset,
  title={Sups: A simulated underground parking scenario dataset for autonomous driving},
  author={Hou, Jiawei and Chen, Qi and Cheng, Yurong and Chen, Guang and Xue, Xiangyang and Zeng, Taiping and Pu, Jian},
  booktitle={2022 IEEE 25th International Conference on Intelligent Transportation Systems (ITSC)},
  pages={2265--2271},
  year={2022},
  organization={IEEE}
}

\end{document}